\pdfoutput=1

\documentclass[11pt]{article}

\usepackage[]{ACL2023}

\usepackage{times}
\usepackage{latexsym}

\usepackage[T1]{fontenc}

\usepackage[utf8]{inputenc}

\usepackage{microtype}
\usepackage{enumitem}
\usepackage{inconsolata}

\usepackage{graphicx}
\graphicspath{ {./images/} }
\usepackage{amsmath}
 \usepackage{multirow}
 \usepackage[labelfont=bf]{caption}
\title{"Let's not Quote out of Context": Unified Vision-Language Pretraining for Context Assisted Image Captioning}

\author{Abisek Rajakumar Kalarani \and Pushpak Bhattacharyya\\
        Department of Computer Science and Engineering, IIT Bombay, India\\
        \texttt{\{abisekrk, pb\}@cse.iitb.ac.in}\\\AND
        Niyati Chhaya \and Sumit Shekhar\\
        Adobe Research, India\\
        \texttt{\{nchhaya, sushekha\}@adobe.com}}

\begin{document}
\maketitle
\begin{abstract}
Well-formed context aware image captions and tags in enterprise content such as marketing material are critical to ensure their brand presence and content recall. Manual creation and updates to ensure the same is non trivial given the scale and the tedium towards this task. We propose a new unified Vision-Language (VL) model based on the One For All (OFA) model, with a focus on context-assisted image captioning where the caption is generated based on both the image and its context. Our approach aims to overcome the context-independent (image and text are treated independently) nature of the existing approaches. We exploit context by pretraining our model with datasets of three tasks- news image captioning where the news article is the context, contextual visual entailment, and keyword extraction from the context. The second pretraining task is a new VL task, and we construct and release two datasets for the task with  $1.1$M and $2.2$K data instances.  Our system achieves state-of-the-art results with an improvement of up to $\mathbf{8.34}$ CIDEr score on the benchmark news image captioning datasets. To the best of our knowledge, ours is the first effort at incorporating contextual information in pretraining the models for the VL tasks.

\end{abstract}

\begin{figure}
    \centering
    \includegraphics[width=\linewidth]{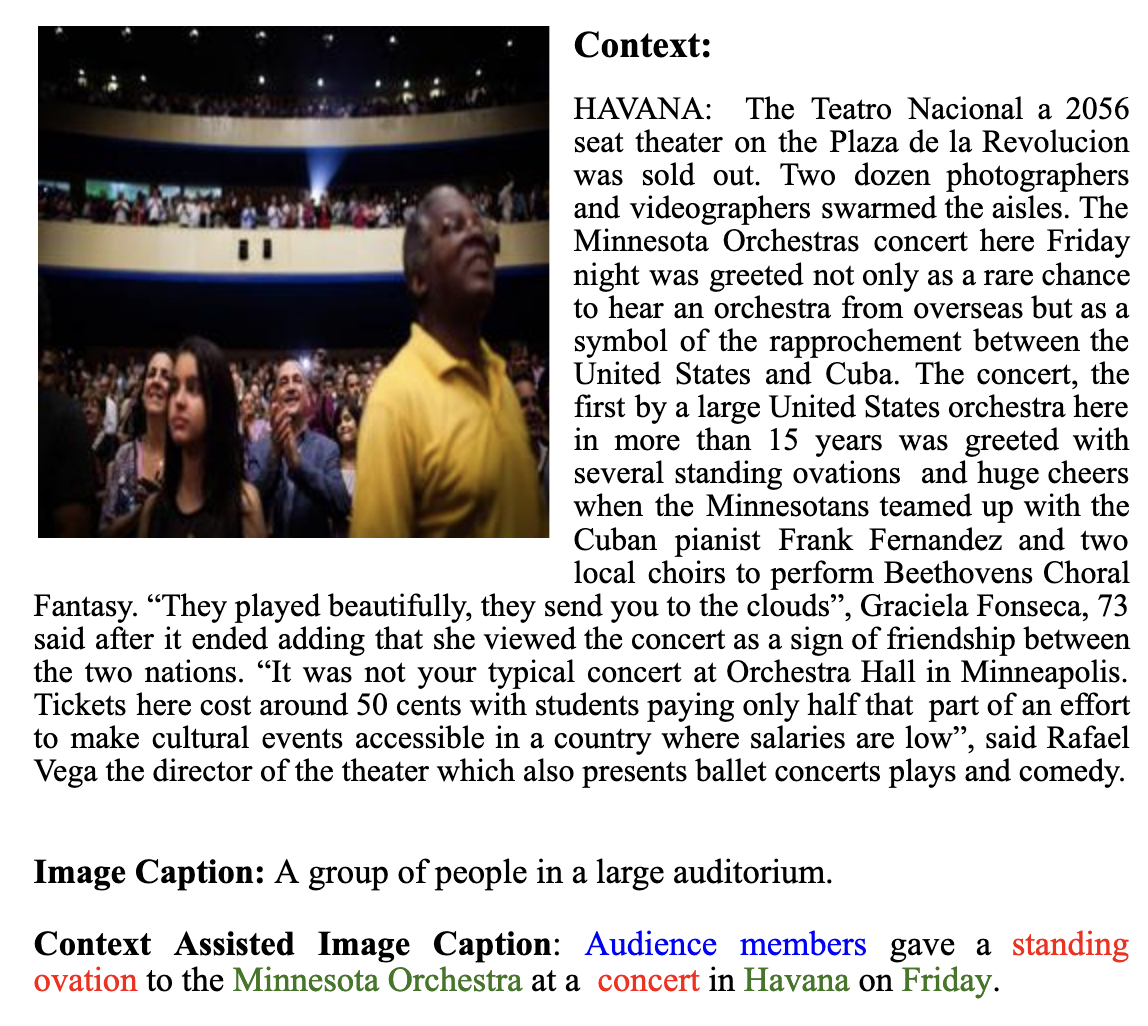}
    \caption{An example image with its context. Text in blue and green can be inferred from the image and the context respectively. Text in red requires both the image and the context.}
    \label{fig: dataset example}
\end{figure}

\section{Introduction}



Large enterprises have several teams to create their content for the purpose of marketing, campaigning, or even maintaining a brand presence. Multimodal assets, particularly images are an integral part of this. The scale and the speed at which one needs to create and update content, especially to ensure personalization requires several resources, which in turn acts as a hindrance to the success of the enterprise. Opportunistic updates are critical for success in the current competitive marketing and advertising scenario. Ensuring that every multimodal asset associated with any piece of enterprise content has an appropriate caption and tag is not possible given the scale. While apparently a nuanced aspect of any image, the caption serves as the key information carrier of what the image is all about, in turn ensuring the right recall (ability to find) for the content that contains this image. If an image has a well-formed caption, that captures the context accurately -- it also makes the document accessible. Making enterprise content accessible is an important metric for large organizations as they strive to be inclusive. The scale and the quick turn-around time demanded for the content creation cycle results in the lack of correct tagging and captions of images (multimodal assets), in turn an eventual lost of revenue and a target customer base. We propose a method towards automated context-aware captioning of images targeted to reduce the tedium and this critical gap in the enterprise authoring and content creation process.

Large-scale pretraining of language models (\citealt{Devlin2019BERTPO}; \citealt{https://doi.org/10.48550/arxiv.2005.14165}) has witnessed great success in many downstream NLP tasks. This success has inspired multi modal pretraining for image-text, image-only, and video-text tasks. Currently, building unified models that jointly learn multiple vision-language tasks is gaining a lot of attention and has shown promising results on many VL tasks (\citealt{Wang2022UnifyingAT}, \citealt{Lu202012in1MV}, \citealt{Cho2021UnifyingVT}, \citealt{Wang2021UFOAU}).

The existing unified Vision-language models focus on  tasks like image captioning \cite{Stefanini2022FromST}, visual question answering \cite{Wu2017VisualQA}, visual entailment \cite{Xie2019VisualEA}, and image-text retrieval \cite{Wang2019LearningTN} that consider the image as a standalone entity. However, images are typically accompanied by text that adds additional meanings which are not utilized in these tasks as shown in Figure \ref{fig: dataset example}. Also, the same image can mean different things in different contexts. For example, a picture of a football player being emotional can mean they are celebrating a goal or are disappointed with their shot, depending on the context. Hence it is essential to consider the context of the image for understanding it completely.

Traditional image captioning models do not use contextual information. In news image captioning \cite{Biten2019GoodNE}, the generated caption contains information extracted from both the news article and the image. The news image captioning task is a special subtask of context assisted image captioning task that uses the news article as the contextual information about the image. In our work, the task names- news image captioning and context assisted image captioning are hence used interchangeably. Existing pretrained VL models lack the ability to use contextual information as the pretraining tasks do not contain long text associated with image-text pairs. We propose a new unified VL model based on the One For All (OFA) model, with a focus on using the contextual information associated with the image for real-world problems like news image captioning.

As there are no existing VL classification task that uses contextual information, we introduce a new VL task called `Contextual Visual Entailment'. Visual entailment \cite{Xie2019VisualEA} is a refined image-text matching task that checks for the entailment of the caption with the premise image. Visual entailment deals with only the descriptive characteristics of the image. 
In our contextual visual entailment task, both the image and the context of the image are treated as the premise, and the entailment of the caption is predicted with respect to both.


Our contributions are:
\begin{itemize}[leftmargin=*,noitemsep]

    \item A new unified VL model pretrained for keyword extraction, contextual visual entailment, and news image captioning with a focus on using contextual information which has not been explored before.


    \item State-of-the-art results on the GoodNews and NYTimes800k datasets with an improvement of $\mathbf{8.34}$ CIDEr points on the GoodNews dataset.


    \item A novel VL classification task where the context information surrounding the image is utilized for detecting the entailment of the caption with the image.


    \item Release of two datasets\footnote{Code and data are available at: \url{https://github.com/abisekrk/context-assisted-image-captioning}}- a large synthetic dataset consisting of $\mathbf{1.1M}$ Image-Caption pairs with context and a more challenging dataset with manually annotated negative samples consisting of $\mathbf{2.2K}$ instances for the proposed contextual visual entailment task.

    

\end{itemize}





\section{Related Work}
\textbf{Image Captioning} was initially conceived as a caption retrieval or template filling task. It involved matching the query image with a predefined set of captions or identifying the objects in the image to place them in predefined templates (\citealt{Farhadi2010EveryPT}; \citealt{li-etal-2011-composing}; \citealt{babytalk}). The advancements made with deep learning based techniques in machine translation inspired the community to adopt similar techniques for image captioning where images were fed to the encoder and the decoder generated caption as a sequence of words (\citealt{Farhadi2010EveryPT}; \citealt{li-etal-2011-composing}; \citealt{babytalk}). Attention allows decoder to focus on different parts of the input and hence it was incorporated to generate words focused on important regions of the image in both sequential models (\citealt{show_attend_tell}; \citealt{adaptive_attention}; \citealt{bottom_up_attention}; \citealt{Huang2019AttentionOA}) and transformer based models \cite{Cornia_2020_CVPR}. In recent years, the models are trained on huge datasets with several millions of image-text pairs for image captioning (\citealt{Li2020OscarOA}; \citealt{Su2020VLBERTPO}; \citealt{Radford2021LearningTV}).
However, in all these works images are treated as standalone entities and their context is not taken into account.

\textbf{News Image Captioning} deals with the generation of captions for news images. The news articles contain the context of the image and they are taken into account during the caption generation process. \citet{Biten2019GoodNE} propose an encoder-decoder model with attention over both image and news article encodings to generate news image captions. It generates captions with placeholders for named entities and fills those placeholders by choosing named entities from the news article. \citet{news_cap_with_sum} model it as a query-based summarization problem where the news image acts as the query and the news article is the source text to be summarized.  \citet{Tran2020TransformAT} use transformers with separate encoders for extracting image features, object features and faces present in the image. The decoder receives the input from all three encoders to generate caption. \citet{liu-etal-2021-visual} use a visual selective layer that learns to align the image features with the text in the news article to generate captions. \citet{yang-etal-2021-journalistic} discuss the journalistic guidelines followed while writing news image captions in journals and incorporate them in the generation process. \citet{10.1145/3503161.3547883} use prompt tuning to finetune pretrained models for news image captioning.

\textbf{Unified Vision-Language (VL) modeling} is a new paradigm that involves creating a unified framework for multiple vision-language tasks, allowing models to be trained on a range of datasets constructed for a range of tasks. ViLBERT \cite{Lu2019ViLBERTPT} extends the BERT \cite{Devlin2019BERTPO} architecture to work with visual inputs. \citet{Lu202012in1MV} propose a multi-task training approach with 12 VL datasets on 4 broad tasks. VL-T5 \cite{Cho2021UnifyingVT} combines multiple VL tasks as text generation tasks using pretrained models for image features. UniT \cite{Hu2021UniTMM} unifies cross-modal tasks by using a modality specific encoder and a shared decoder. UFO \cite{Wang2021UFOAU} proposes to use the same transformer architecture as the encoder for both image and text in VL tasks. UniTAB \cite{Yang2021CrossingTF} supports VL tasks with bounding boxes by encoding the text and box output sequences to shared token sequences. OFA \cite{Wang2022UnifyingAT} abstracts all VL tasks into sequence-to-sequence problems.


Existing unified VL models do not consider the context of the image in their pretraining tasks. Our unified model is pretrained with tasks that include contextual information and hence it achieves state-of-the-art results on news image captioning datasets.

\begin{figure*}[h]
    \centering
    \includegraphics[height=7cm, keepaspectratio]{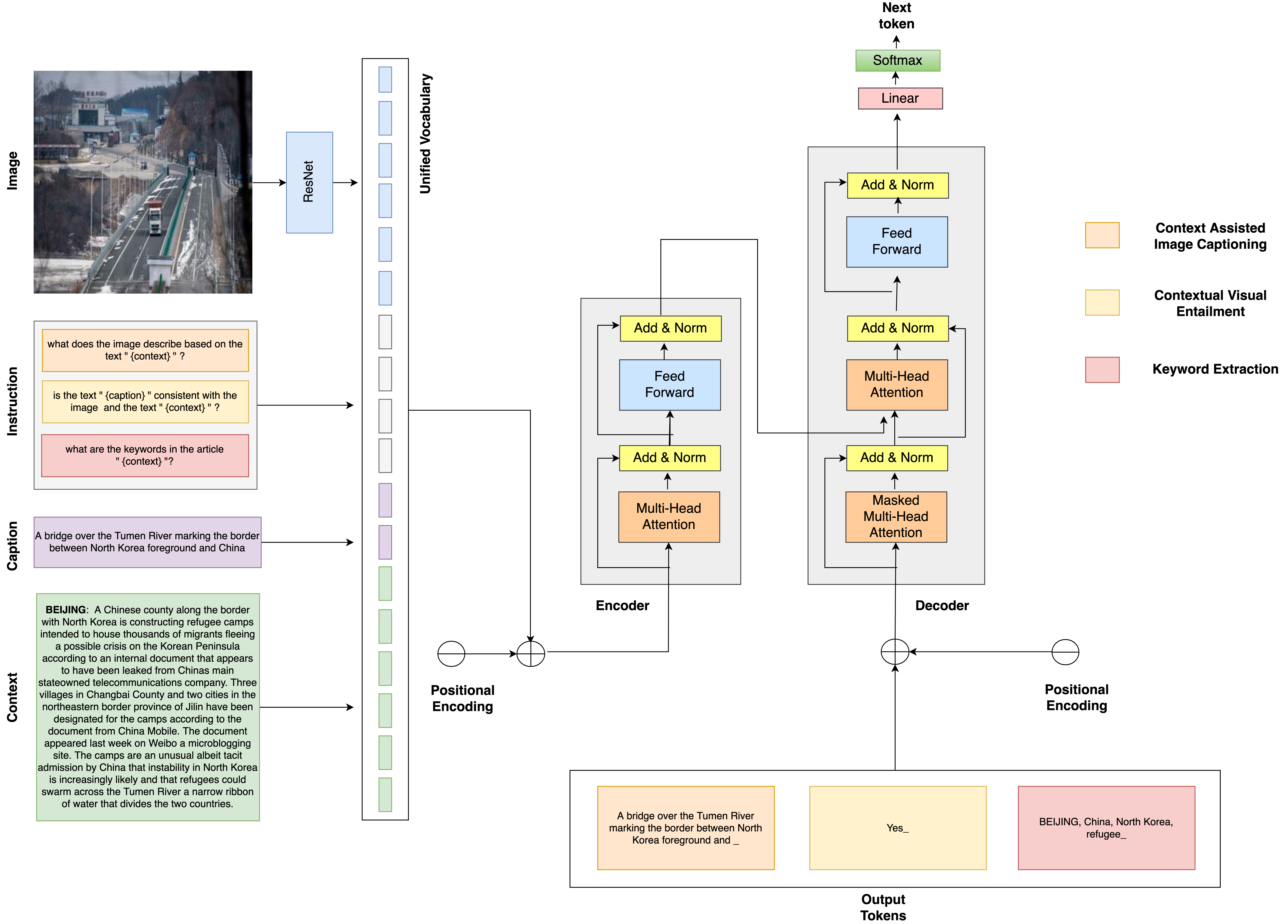}
    \caption{An overview of our unified Vision-Language model pretrained for the three subtasks- context assisted image captioning, contextual visual entailment, and keyword extraction.}
    \label{fig: OFA architecture diagram}
\end{figure*}

\section{Dataset} \label{section: dataset}
We pretrain our model on a large pretraining dataset and evaluate its performance on benchmark datasets for news image captioning.
\subsection{Pretraining Datasets}
We use Visual News \cite{liu-etal-2021-visual} and KPTimes \cite{gallina2019kptimes} datasets for constructing our pretraining datasets. 

Visual News dataset was compiled by collecting news articles from four news agencies: The Guardian, BBC, USA Today, and The Washington Post. It only includes the articles with high resolution images and where the caption length is between $5$ and $31$ words. It is diverse with differences in properties like average caption length, article length, and distribution of named entities across news agencies.

KPTimes dataset was constructed by crawling over 0.5  million news articles, mainly from New York Times. The metadata associated with field- "news\_keywords" and "keywords" form the gold standard keyphrases . The three pretraining datasets are constructed from these datasets.

\textbf{News Image Captioning:} We removed duplicate captions, and news articles without images, and captions from Visual News dataset. The cleaned dataset consists of $11,97,000$ data instances with image, news article, and the caption. These are split as $11,17,697$ for training, $40,000$ for validation, and $40,000$ for test sets respectively.

\textbf{Contextual Visual Entailment} is a binary classification problem, so it is required to construct both positive and negative pairing of the image, and caption with the context. For the data instances where the caption entails the image and the context, the original image, the caption, and the context from the training split of the above news image captioning dataset are used {(P)}. We use the following operations to generate the inconsistent pairs in our dataset:
\begin{enumerate}[leftmargin=*,noitemsep]
    \item Choose a random caption different from the correct caption \textbf{(N-I)}.
    \item Replace the named entities in the correct caption with named entities from randomly chosen caption \textbf{(N-II)}. For example, the caption \emph{`John Garrison performing in Berlin, April 2015'} will be changed to \emph{`Mark Pattinson performing in London, April 2015'}.
    \item Keep the named entities of the original caption intact but replace the remaining content with a random caption that has the same type and the same number of named entities \textbf{(N-III)}. For example, the caption \emph{`John Garrison performing in Berlin, April 2015'} will be changed to \emph{`John Garrison waiting in queue for filing tax returns in Berlin, April 2015'}.
\end{enumerate}
Named entity recognition is done with SpaCy \cite{spacy2} in our experiments. SpaCy allows the detection of $18$ different named entities. We only use the named entities labeled as \textit{`PERSON', `FAC', `ORG', `GPE', `LOC',} and \textit{`EVENT'} that represent a person, building/airport, organization, geopolitical entities, location, and event respectively, as they occur more frequently.

The N-I class of negative captions will have different information and different named entities from the original caption. The N-II class will have same information as the original caption but will contain different named entities. The N-III class of captions will have same named entities but will convey different information. The final dataset has $1005925$, $55884$, and $55884$ instances in the train, validation and test split respectively. We also create a separate manually annotated challenging dataset for evaluation.

In addition to synthetically creating a dataset for pretraining, we create and release a manually annotated challenging dataset for the task of contextual visual entailment consisting of $2.2K$ data instances. The negative captions in this dataset are created manually by  changing a word or a small phrase from the original caption, such that its meaning changes significantly without much difference in the sentence structure. For example, \emph{`Supporters marched peacefully during the protest'} will be changed to \emph{`Supporters marched violently during the protest'}. The negative examples created in these ways will ensure that the models need to learn the relationship between image, caption, and context to identify the entailment correctly. This is used to test the model's knowledge of image-caption entailment at a more finer level. 

\textbf{Annotation Details:} The annotations were performed by two annotators proficient in English. One is a master's student and the other is a bachelor's student in Computer Science and Engineering. They were provided with examples of negative captions before annotation. The image links, caption, and context were shared in Google Sheets for annotation. The annotators were only asked to select a word or phrase from the caption and replace it with a new word or phrase. The modified captions were exchanged and verified by each other.


\begin{table*}
\centering
\begin{tabular}{|c|l|c|c|c|c|c|c|} 

\hline
\multirow{2}{*}{\textbf{Dataset}} & \multirow{2}{*}{\textbf{Model}} & \multirow{2}{*}{\textbf{B-4}} & \multirow{2}{*}{\textbf{MET.}} & \multirow{2}{*}{\textbf{ROUGE}} & \multirow{2}{*}{\textbf{CIDEr}} & \multicolumn{2}{c|}{\textbf{Named Entities}}   \\ 
\cline{7-8}
                                  &                                 &                                  &                                  &                                 &                                 & \textbf{P}            & \textbf{R}             \\ 
\hline
\multirow{7}{*}{\textbf{GoodNews}}         & GoodNews                        & 1.86                             & 13.75                            & 20.46                           & 17.57                           & 8.23                  & 6.06                   \\ 
\cline{2-8}
                                  & Transform and Tell              & 6.05                             & 10.30                            & 21.40                           & 54.30                           & 22.20                 & 18.70                  \\ 
\cline{2-8}
                                  & Visual News                     & 6.10                             & 8.30                             & 21.60                           & 55.40                           & 22.90                 & 19.30                  \\ 
\cline{2-8}
                                  & JoGANIC                         & 6.83                             & 11.25                            & 23.05                           & 61.22                           & 26.87                 & 22.05                  \\ 
\cline{2-8}
                                  & NewsMEP                         & 8.30                             & 12.23                            & 23.17                           & 63.99                           & 23.43                 & 23.24                  \\ 
\cline{2-8}
                                  & OFA                             & 6.41                             & 10.63                            & 23.59                           & 67.19                           & 23.06                 & 19.04                  \\ 
\cline{2-8}
                                  & \textbf{Ours}                   & 7.14                             & 11.21                            & \textbf{24.30}                  & \textbf{72.33}                  & 24.37                 & 20.09                  \\ 
\hline
\hline
\multirow{6}{*}{\textbf{NYTimes800K}}      & Transform and Tell              & 6.30                             & 10.30                            & 21.70                           & 54.40                           & 24.60                 & 22.20                  \\ 
\cline{2-8}
                                  & Visual News                     & 6.40                             & 8.10                             & 21.90                           & 56.10                           & 24.80                 & 22.30                  \\ 
\cline{2-8}
                                  & JoGANIC                         & 6.79                             & 10.93                            & 22.80                           & 59.42                           & 28.63                 & 24.49                  \\ 
\cline{2-8}
                                  & NewsMEP                         & 9.57                             & 13.02                            & 23.62                           & 65.85                           & 26.61                 & 28.57                  \\ 
\cline{2-8}
                                  & OFA                             & 6.91            & 10.77            & 22.70           & 61.81           & 27.14 & 22.51  \\ 
\cline{2-8}
                                  & \textbf{Ours}                            & 7.54                             & 11.27            & 23.28                           & \textbf{66.41}                           & 28.11                 & 23.25                  \\
\hline
\end{tabular}
\caption{Experimental results on the GoodNews and NYTimes800K datasets compared with other models. P and R denote the precision and recall of generating named entities. B-4 indicates BLEU-4 and MET. indicates METEOR.}
\label{table: result comparison}
\vspace{-0.5cm}
\end{table*}

\subsubsection{Keyword Extraction}
The dataset for the keyword extraction task is constructed from KPTimes dataset after removing duplicate news articles. The news article forms the input to the system and the sequence of keywords form the output. The final dataset has $259902$, $10000$, and $10000$ data instances in the train, validation and test split respectively. The training data from these three datasets are combined to generate the pretraining dataset with 2.3M data instances.

\subsection{Benchmark Datasets}
The performance of models trained on the pretraining datasets is evaluated on two benchmark datasets- GoodNews \cite{Biten2019GoodNE} and NYTimes800K \cite{Tran2020TransformAT}. We follow the train, validation, and test splits from the original work for both datasets. The GoodNews dataset has $424, 000$, $18, 000$, and $23, 000$ in training, validation, and test split respectively.  The NYTimes800K dataset has $763,000$ training, $8000$ validation, and $22,000$ test instances in the dataset.

\section{Our Model}

Unified Vision-Language (VL) modeling has shown great promise in multiple VL tasks. Hence, we use a unified model for all three tasks- context assisted image captioning, contextual visual entailment, and keyword extraction. Figure \ref{fig: OFA architecture diagram} shows an overview of our unified VL pretraining strategy. We use the OFA$_{Large}$ \cite{Wang2022UnifyingAT} architecture. OFA is a task and modality agnostic  model that unifies all vision-language, vision-only, and language-only tasks using a sequence-to-sequence learning framework. We use ResNet152 \cite{He2016DeepRL} and VQGAN \cite{Esser2020TamingTF} to obtain visual tokens for the given image. The text (context and caption) is tokenized by byte-pair encoding (BPE). A single unified vocabulary is used for both visual and linguistic tokens. Transformers are used as encoders and decoders and all vision-language tasks are abstracted to seq-to-seq conversion tasks with specific instructions created for each task, similar to the OFA pretraining.

The pretraining of our model involves three tasks- News image captioning, contextual visual entailment, and keyword generation. For news image captioning, we convert the image, caption, and context into a sequence of input tokens and generate the caption as a sequence of tokens conditioned on these input tokens. For keyword generation, the news article is tokenized as the input sequence and the keywords are generated by the model as the output sequence. Contextual visual entailment is a classification task, so the input sequence to the model is the image, caption, and context tokens and the model is trained to generate `Yes' or `No' as the output indicating if the caption entails the image and context or not respectively.

The model is trained to reduce the cross entropy loss. For the input sequence x consisting of visual and text tokens and output y, the loss is given as:
$$
    \mathcal{L} = \sum_{i}^{|y|}logP_{\theta}(y_i| y_{<i}, x)
$$
where $y_i$ is the text token to be predicted and $y_{i-1}$, $y_{i-2}$ are tokens predicted so far.

\begin{table*}
\centering
\begin{tabular}{|l|c|c|c|c|} 
\hline
\textbf{Model}        & \textbf{BLEU-4} & \textbf{METEOR} & \textbf{ROUGE} & \textbf{CIDEr}  \\ 
\hline
BLIP-2 + GPT-3        & 2.06            & 8.48            & 13.22          & 17.12           \\ 
\hline
OFA                   & 6.41            & 10.63           & 23.59          & 67.19           \\ 
\hline
OFA + Captioning + Contextual Visual Entailment        & 6.90            & 11.01           & 23.83          & 69.97           \\
\hline
OFA + Captioning + Keyword Extraction        &    6.69        &  10.81          &   23.44        &   67.83         \\
\hline
OFA + Captioning        &  6.85           &     10.90       &   23.70       &        68.93    \\
\hline
Our Model (Without Context) & 2.24            & 5.34            & 14.45          & 18.04           \\ 
\hline
Our Model + NE        & 6.95            & 11.06           & 23.98          & 70.03           \\

\hline
Our Model             & \textbf{7.14}            & \textbf{11.21}           & \textbf{24.30}          & \textbf{72.33}           \\

\hline
\end{tabular}
\caption{Ablation study results on the GoodNews dataset. NE denotes fine-tuning done with named entities extracted separately. `OFA + X' denotes pretraining of OFA done with X task. `Our Model' refers to the model pretrained on the three tasks with context information.}
\label{table: ablation}
\vspace{-0.3cm}
\end{table*}

\section{Experiments}
The experimental details for pretraining and finetuning for context assisted image captioning are discussed here.
\subsection{Pretraining}
We used OFA$_{Large}$ architecture for pretraining our model. The model has 472M parameters with 12 encoder and 12 decoder layers. The weights were initialized with the publicly available OFA$_{Large}$ checkpoint to retain the knowledge from other VL tasks. The model was pretrained on the 2.3M data instances from the pretraining datasets.

All 3 tasks were abstracted into sequence-to-sequence task. For the instances of news image captioning dataset, the instruction was ``\emph{What does the image describe based on the text  <context> ?}", where <context> holds the tokens from the news article. For contextual visual entailment, the instruction was ``\emph{Is the text <caption> consistent with the image  and the text <context> ?}", where <caption> and <context> contain the text tokens from the caption and the context. For the keyword extraction task, the instruction given was ``\emph{What are the keywords in the article <context>?}".
\subsection{Context Assisted Image Captioning}
Our unified model pretrained model for the three tasks was finetuned on GoodNews and NYTimes800K datasets for the task of context assisted image captioning. The image resolution was fixed at $384 * 384$ and the news article was clipped to $512$ tokens. The maximum caption length was fixed at $30$. We use a batch size of $8$ for training. We train the model with early stopping and choose the model that achieves the best CIDEr score on the validation set. The best-performing model is then tested on the unseen test data and the results are summarized in Table \ref{table: result comparison}. 




\subsection{Training Details} \label{training details}
The experiments were done with the OFA$_{Large}$ architecture. For both pretraining and finetuning, the image resolution was fixed at $384*384$. The input token length was restricted to $512$ tokens while the output was restricted to $30$ tokens. The dropout ratio was set to 0.1. We used Adam optimizer \cite{Kingma2014AdamAM} with $0.9$ and $0.999$ as the $\beta$ values with $\epsilon = 1e-08$ and warm-up ratio was set as $0.06$. We used an initial learning rate of $1e-5$ with polynomial decay. We used a beam size of $10$ during the test inference with temperature $0.98$. We also used mixed precision training to speed up the training process.

\subsection{Frozen Image Encoder + Frozen LLM}
LLMs and large-scale pretrained VL models have shown great zero-shot performance in many downstream applications. We use BLIP-2 \cite{Li2023BLIP2BL} for getting zero-shot image captions for the GoodNews dataset. These captions are generated without contextual information and are descriptive in nature. These captions are passed to a LLM along with contextual information to generate context assisted image captions. We use \texttt{text-davinci-003} model in the GPT-3 family \cite{Ouyang2022TrainingLM}. The prompt for generating the caption was "Add contextual information to the caption. Caption: <sample caption> Context: <sample context>". We randomly sampled a caption, and context pair from the training dataset of the GoodNews dataset and used it as an example in the prompt. The contextual captions are predicted for caption and context pair in the test set. The results are discussed in Table \ref{table: ablation}.
\subsection{Ablation Study}
In order to analayse the importance of the three pretraining tasks we used, we pretrained the OFA model using three different subsets of the pretraining tasks. We pretrained the model with only ``Captioning and Contextual Visual Entailment" tasks, with only ``Captioning and Keyword Extraction" task and with only ``Captioning" task and compared their performance with the model trained on all the three tasks.

Previous works in news image captioning (\citealt{liu-etal-2021-visual}; \citealt{yang-etal-2021-journalistic}; \citealt{10.1145/3503161.3547883}) have shown that extracting named entities from the context and feeding them to the decoder helps generate correct named entities in the caption. Hence, we also try injecting named entities into the prompt while finetuning the model. We use SpaCy for identifying and extracting named entities. We update the prompt as "what does the image describe about the names <named entities> based on the text <context>?" during finetuning. We clipped the named entity tokens to $64$ and restricted the context to $512$ tokens as done in previous experiments. We also perform experiments without using the contextual information with our pretrained model in the traditional image captioning setting, to analyze the usefulness of the contextual information. The results are summarized in Table \ref{table: ablation}.


\section{Results and Analysis}
Our model achieves state-of-the-art results on both GoodNews and NYTimes800K datasets. The OFA model finetuned on benchmark datasets also shows good performance. This shows the ability of OFA to adapt to new tasks and the correctness of our instructions for finetuning. However, it can be seen that due to the lack context information in the pretraining tasks used in OFA, the model doesn't produce substantially better results compared to the current SOTA models. Our model pretrained on the three tasks shows a $5.14$ CIDEr score improvement over the OFA model on the GoodNews dataset which is an $8.34$ CIDEr score improvement over the current SOTA model. The model also achieves a SOTA result of $66.41$ CIDEr score on the NYTimes800K dataset. 

The average length of news articles in GoodNews and NYTimes800K dataset are 451 and 974 words respectively. The larger article length in NYTimes800K dataset is the reason for the CIDEr scores being closer to the current SOTA as the context length in  our experiments is restricted to 512 tokens. We also obtain comparable performance in precision and recall of named entity generation despite not feeding the named entities directly to the model like in the previous works.

The BLIP-2 model has shown great promise in zero-shot image caption generation. We use BLIP-2 to generate descriptive captions and feed those captions as input to the GPT-3 model along with the context to generate the final context assisted caption. The BLIP-2 + GPT-3 model generates fluent captions but it does not contain the relevant information based on the image features as indicated by the poor performance on evaluation metrics in Table \ref{table: ablation}. This indicates that it is essential to train with both image and context together.

Our pretraining tasks indirectly direct the model to capture named entity information from the context. However, earlier works on news image captioning show that extracting named entities and feeding them directly to the model can help it generate better captions with correct named entities. Our pretrained model showed a slight decrease in performance when named entity information is presented to it in the prompt. This is because the named entity tokens take up valuable space in the 512 tokens allowed for the context, leading to information loss gained from the context. Also, since only 64 tokens are allowed for named entities, not all the important entities in the news article are presented in the prompt and it disadvantages the model in both ways.

We also pretrained the OFA model with a subset of the three pretraining tasks to identify the importance of each task in pretraining. The models pretrained with a combination of two tasks and with only the captioning task performed poorly compared to the model pretrained on the three tasks. This shows the importance of training with all three tasks. Between the two two-task pretrained models, the model that used contextual visual entail task performed better, indicating the usefulness of the task we introduced.

\section{Summary and Conclusion}
In this work, we proposed a new unified VL model that uses contextual information of images that has not been utilized in pretraining before. We introduce a new VL classification task called contextual visual entailment and pretrain a model with three subtasks that uses long text along with image and caption. Our model achieves new state-of-the-results on benchmark datasets for news image captioning and highlights the importance of using contextual information in pretraining.

In the future, we aim to deploy our model to allow context-aware caption generation which could be used in enterprise authoring and many other content creation processes.





\bibliography{anthology,custom}
\bibliographystyle{acl_natbib}

\appendix

\section{Appendix}
\label{sec:appendix}
\begin{table*}[h]
\centering
\begin{tabular}{|l|l|r|r|r|} 
\hline
\multicolumn{2}{|l|}{}                                               & \multicolumn{1}{l|}{\textbf{Train}} & \multicolumn{1}{l|}{\textbf{Val}} & \multicolumn{1}{l|}{\textbf{Test}}  \\ 
\hline
\multirow{3}{*}{\textbf{Pretraining}} & Keyword Extraction           & 259902                              & 10000                             & 10000                               \\ 
\cline{2-5}
                                      & Contextual Visual Entailment & 1005925                             & 55884                             & 55884                               \\ 
\cline{2-5}
                                      & News Image Captioning        & 1117697                             & 40000                             & 40000                               \\ 
\hline
\multirow{2}{*}{\textbf{Benchmark}}   & GoodNews                     & 424000                              & 18000                             & 23000                               \\ 
\cline{2-5}
                                      & NyTimes800K                  & 763000                              & 8000                              & 22000                               \\
\hline
\end{tabular}
\caption{Summary of the statistics of the datasets used for pretraining and benchmarking.}
\label{table: dataset stats}
\end{table*}

\subsection{Dataset Details}


Table \ref{table: dataset stats} provides the summary of the three datasets used for pretraining.

\begin{table*}
\centering
\begin{tabular}{|l|l|r|r|r|r|} 
\hline
\multicolumn{2}{|c|}{\multirow{2}{*}{\textbf{Model}}}         & \multicolumn{4}{c|}{\textbf{Overall}}                                                                                                            \\ 
\cline{3-6}
\multicolumn{2}{|c|}{}                                        & \multicolumn{1}{c|}{\textbf{Acc.}} & \multicolumn{1}{c|}{\textbf{Pre.}} & \multicolumn{1}{c|}{\textbf{Rec.}} & \multicolumn{1}{c|}{\textbf{F1}}  \\ 
\hline
\multirow{3}{*}{\textbf{w/o context}} & 1) CLIP + FNN         & 61.30                              & 61.61                              & 60.00                              & 60.79                             \\ 
\cline{2-6}
                                      & 2) CLIP + Transformer & 60.00                              & 59.83                              & 60.87                              & 60.34                             \\ 
\cline{2-6}
                                      & 3) Ours               & 66.96                              & 69.70                              & 60.00                              & 64.49                             \\ 
\hline
\multirow{3}{*}{\textbf{w/o image}}   & 4) CLIP + FNN         & 59.13                              & 61.54                              & 48.70                              & 54.37                             \\ 
\cline{2-6}
                                      & 5) CLIP + Transformer & 56.96                              & 56.25                              & 62.61                              & 59.26                             \\ 
\cline{2-6}
                                      & 6) Ours               & 65.22                              & 66.67                              & 60.87                              & 63.64                             \\ 
\hline
\multirow{3}{*}{\textbf{w/ context}}  & 7) CLIP + FNN         & 64.35                              & 63.64                              & 66.96                              & 65.25                             \\ 
\cline{2-6}
                                      & 8) CLIP + Transformer & 65.65                              & 63.64                              & \textbf{73.04}                              & 68.02                             \\ 
\cline{2-6}
                                      & 9) Ours               & \textbf{73.04}                              & \textbf{79.78}                              & 61.74                              & \textbf{69.61}                             \\
\hline
\end{tabular}
\caption{Experimental results on the manually annotated contextual visual entailment dataset, where w/o context, w/o image and w/ context indicate experiments done without context (Image + Caption), without image (Caption + Context), and with context (Image + Caption + Context).}
\label{table: evaluation results on CD}
\end{table*}



\subsection{Additional Experiments}
Our pretrained model achieves state of the results on news image captioning task. In addition, it performs very well on the other two pretraining tasks. The contextual visual entailment is a new task introuduced by our work and hence we propose baselines for comparing the results of our model. We compare our model's performance on keyword extraction against standard works.

\subsubsection{Contextual Visual Entailment}

We propose a two baselines for contextual visual entailment, where the image and text features are extracted from pretrained networks. The features are obtained from a pretrained CLIP (Contrastive Language–Image Pre-training)  model. CLIP \cite{Radford2021LearningTV} was trained on large scale image-text corpus to minimize contrastive loss such that the text embedding and the image embedding will have higher cosine similarity if the text describes the image perfectly and low when the text incorrectly describes the image.

\subsubsection*{CLIP and FNN model}
In our CLIP embedding based models, the representation for image, caption, and context is obtained from a pretrained CLIP model. 

The CLIP and FNN model, uses a simple early fusion strategy in which the image, caption, and context embeddings from CLIP are concatenated and fused with feed-forward neural networks. The three input embeddings are concatenated and passed to a two-layer feedforward neural network for combining the information. An output layer predicts the entailment label. The experiments are repeated without the context information and then again without the image features as input to study their impact on entailment detection.


\subsubsection*{CLIP and Transformer model}

The fusion of image-text information using transformers has helped achieve good performance on many standard vision-language tasks (\citealt{ZhouPZHCG20}; \citealt{Chen2020UNITERUI}; \citealt{Wang2022UnifyingAT} ). The CLIP and Transformer model uses transformer \citet{Vaswani2017AttentionIA} encoders with multi-head attention to combine these multimodal information. A transformer layer receives the input features from the image, caption, and context and generates the combined representation for the information. The outputs from the transformer layers are pooled to a single dense layer followed by a classification layer to perform the final binary classification.

We summarize the results of our experiments on the manually annotated contextual visual entailment data in Table \ref{table: evaluation results on CD} and compare it with the results produced by our pretrained model.

\begin{table}
\centering
\begin{tabular}{|l|r|} 
\hline
\multicolumn{1}{|c|}{\textbf{Model}} & \multicolumn{1}{c|}{\textbf{F@10}}  \\ 
\hline
FirstPhrases                         & 9.2                                 \\ 
\hline
MultipartiteRank                     & 11.2                                \\ 
\hline
CopySci                              & 11                                  \\ 
\hline
CopyNews                             & 39.3                                \\ 
\hline
\textbf{Ours}                        & \textbf{40.6}                       \\
\hline
\end{tabular}
\caption{Results of Keyword extraction task on KPTimes dataset. F@10 represents the F1 score at the top N = 10 keyphrases}
\label{table: keyword results}
\end{table}

\subsubsection{Keyword Extraction}
We use our pretrained model to perform keyword extraction as a sequence to sequence task where the output is the set of keyword tokens. We use similar hyperparameters to caption generation for keyword generation. We finetune the pretrained model for $5$ epochs and report the results on KPTimes dataset in Table \ref{table: keyword results}.

\subsection{Examples}

\begin{figure*}
    \centering
    \includegraphics[width=\textwidth,height=\textheight,keepaspectratio]{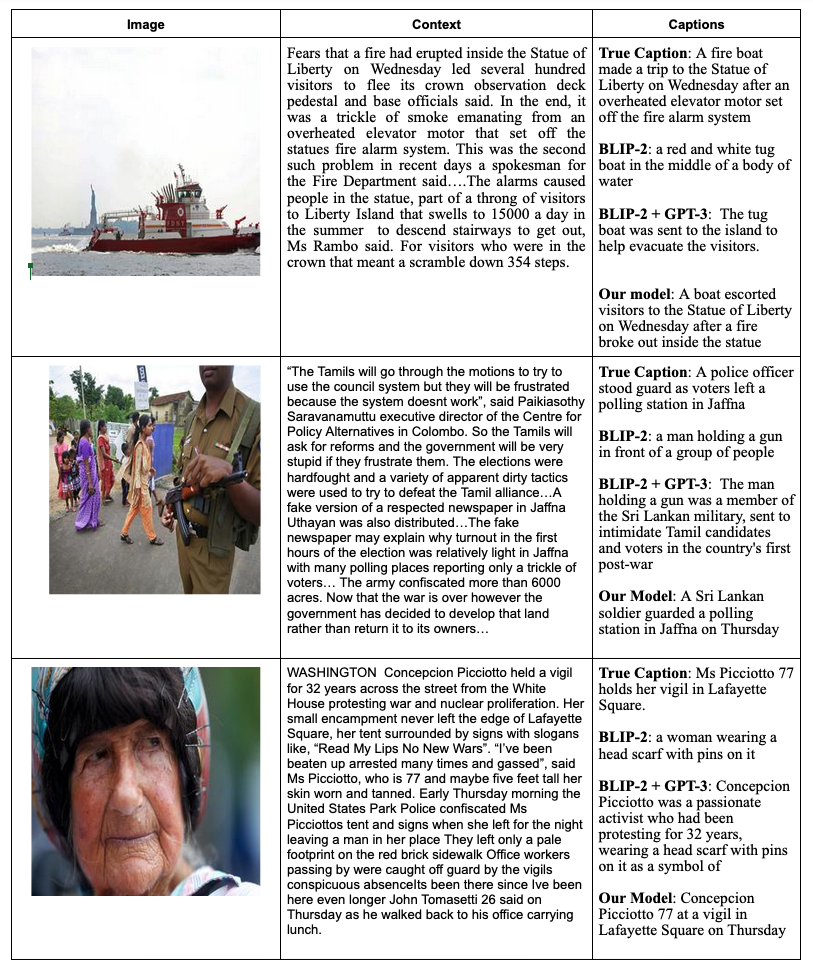}
    \caption{Some examples from our dataset along with the captions generated by each model.}
    \label{fig: appendix_example}

\end{figure*}

Figure \ref{fig: appendix_example} shows examples from the GoodNews dataset and compares the caption generated by each model.

\end{document}